\def\year{2020}\relax
\documentclass[letterpaper]{article}
\usepackage{aaai20}
\usepackage{times}
\usepackage{helvet}
\usepackage{courier}
\usepackage{url}
\usepackage{graphicx}
\usepackage{multirow}
\usepackage{amsmath}
\title{Filling Conversation Ellipsis for Better Social Dialog Understanding}
\author{Xiyuan Zhang,\textsuperscript{\rm 1}
Chengxi Li,\textsuperscript{\rm 1}
Dian Yu, \textsuperscript{\rm 2}
Samuel Davidson, \textsuperscript{\rm 2}
Zhou Yu \textsuperscript{\rm 2}
\\ \textsuperscript{\rm 1}Zhejiang University, 
\textsuperscript{\rm 2} University of California, Davis \\
\{zhangxiyuan, chengxili\}@zju.edu.cn, \{diayu, ssdavidson, joyu\}@ucdavis.edu 
}
\begin{document}
\maketitle
\begin{abstract}
 The phenomenon of ellipsis is prevalent in social conversations. Ellipsis increases the difficulty of a series of downstream language understanding tasks, such as dialog act prediction and semantic role labeling.
We propose to resolve ellipsis through automatic sentence completion to improve language understanding. However, automatic ellipsis completion can result in output which does not accurately reflect user intent. To address this issue, we propose a method which considers both the original utterance that has ellipsis and the automatically completed utterance in dialog act and semantic role labeling tasks. Specifically, we first complete user utterances to resolve ellipsis using an end-to-end pointer network model. We then train a prediction model using both utterances containing ellipsis and our automatically completed utterances. Finally, we combine the prediction results from these two utterances using a selection model that is guided by expert knowledge. Our approach improves dialog act prediction and semantic role labeling by 1.3\% and 2.5\% in F1 score respectively in social conversations. We also present an open-domain human-machine conversation dataset with manually completed user utterances and annotated semantic role labeling after manual completion. 
\end{abstract}

\section{Introduction}
Ellipsis, in which a speaker omits words that are understood from context, is a frequent phenomenon in human conversation. Although natural to humans, ellipsis poses a challenge for language understanding in spoken dialog systems. We find that among 2,000 sample utterances in the Alexa Prize social conversations, about 50\% of the utterances contain some degree of ellipsis.
While humans are generally able to resolve elided elements from context, it is difficult for chatbots to do the same.
Ellipsis can negatively impact the accuracy of language understanding in deployable social chatbots. For example, Table \ref{tab:elli_example} shows an example of a user utterance with ellipsis. It is difficult to tell whether ``what's up with that scene at the end" is a question or an answer to the previous question. However, if we complete the utterance considering the context, we obtain ``I would ask what's up with that scene at the end". It is then easy to understand that the user is stating their opinion with respect to the system's previous question instead of asking a new question.
\begin{table}[h]

    \resizebox{0.95\columnwidth}{!}{
    \begin{tabular}{c|c|c}
    \hline
        \multirow{2}*{\textbf{System Utterance}} &
        \multirow{2}{22mm}{\textbf{User Response\\(original)}}&
        \multirow{2}{22mm}{\textbf{User Response\\(completed)}}\\
         & &\\
    \hline
        \multirow{6}{28mm}{If you got the chance to ask the director of that movie one question, what would it be?}&
        \multirow{6}{22mm}{What's up with that scene at the end.}&
        \multirow{6}{22mm}{{\textit{I would ask}} what's up with that scene at the end.}\\
         & & \\
         & & \\
         & & \\
         & & \\
         & & \\
    \hline
        \multirow{4}{28mm}{Have you read any other books by the same author?}&
        \multirow{4}{22mm}{Okay. (Let's change conversation.)}&
        \multirow{4}{22mm}{Okay {\textit{I have read any other books by the same author.}}}\\
         & &\\
         & &\\
         & &\\
    \hline
    \end{tabular}}
    \caption{Examples of original user utterance and automatically completed utterance. Italics represents the automatically completed portion. The utterance in parentheses is the user utterance in the next turn.}
    \label{tab:elli_example}
\end{table}

A possible way to resolve semantic ambiguity caused by ellipsis is to train a model that can automatically complete sentences with ellipsis. However, automatic completion may introduce errors that can lead to other misunderstandings in downstream tasks. For example, automatically completed utterances might repeat or miss some words. Automatically completed utterances may even result in nonsensical sentences. As shown in Table \ref{tab:elli_example}, the user says ``okay" and pauses before saying ``let's change conversation". Due to an ASR issue, the system ends the user's turn during the pause. However, our automatic completion model might complete the original ``okay" to be ``okay I have read any other books by the same author", which misleads the system that the user is expressing agreement. To mitigate the impact of such completion errors, we propose a hybrid framework that considers both utterances with ellipsis and their automatically completed counterparts, Hybrid-ELlipsis-CoMPlete (Hybrid-EL-CMP),
to improve language understanding. We evaluate the performance on two specific tasks: dialog act prediction and semantic role labeling. We believe other understanding tasks such as syntactic and semantic parsing could also leverage this framework. Hybrid-EL-CMP outperforms models which consider only the original utterances or the automatically completed utterances, respectively.

\begin{figure*}
\centering
\includegraphics[width=2.0 \columnwidth]{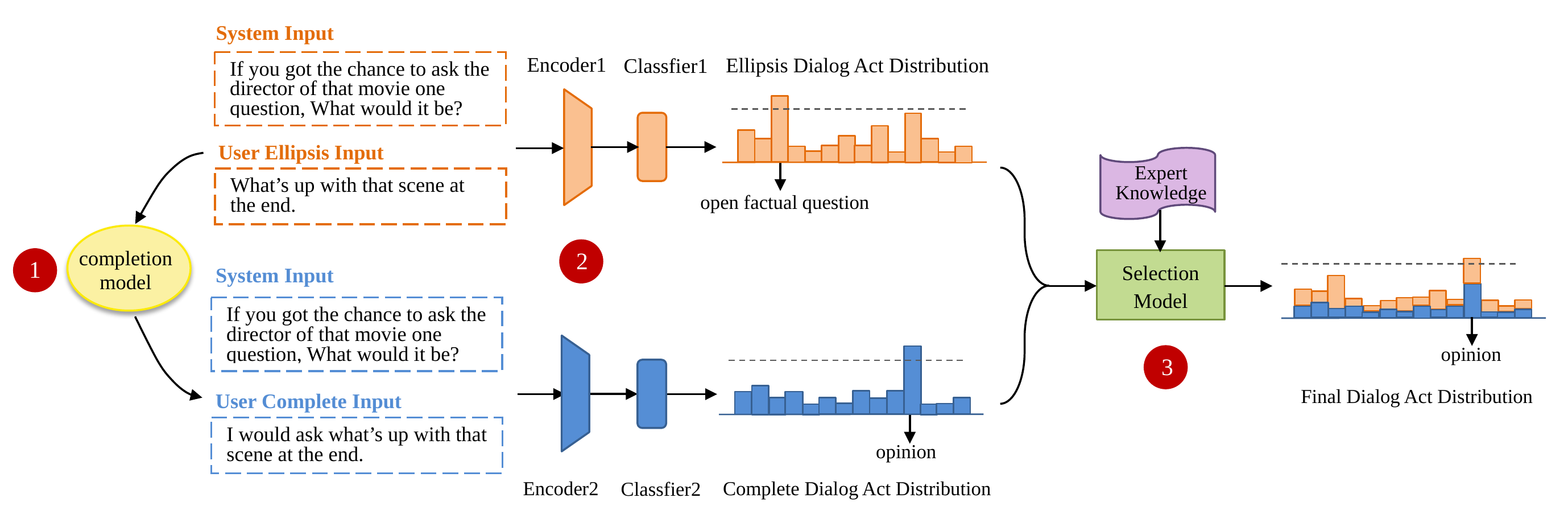}
\caption{Architecture of Hybrid-EL-CMP. Red circles numbered 1 to 3 represent three model components. The dotted line in the distribution represents the threshold for multi-label dialog act prediction.}
\label{fig:1}
\end{figure*}

Hybrid-EL-CMP contains three primary components: a completion model, two encoder-classifier models that separately capture information from original utterances and auto-completed utterances, and a learning-based selection model guided by expert knowledge that combines the results of the two models.
To obtain automatically completed utterances, we train a generative end-to-end completion model leveraging the idea of Pointer Generator \cite{see2017get}. The completed utterance is generated by copying words either from dialogue history or the current utterance as indicated by the copy mechanism. In summary, our main contributions are:
\begin{itemize}
    \item We propose Hybrid-EL-CMP, a framework to jointly utilize utterances with ellipsis and utterances after automatic completion to achieve better performance in dialog understanding tasks. We show that Hybrid-EL-CMP outperforms state-of-the-art methods on dialog act prediction and semantic role labeling tasks in social conversations.
    \item We present an open-domain human-machine conversation dataset with manually completed user utterances. We also annotate semantic roles in this dataset after manual sentence completion. The annotated dataset is publicly available \footnote{https://gitlab.com/ucdavisnlp/filling-conversation-ellipsis}.
\end{itemize}

\section{Related Work}
Our work to improve social dialog understanding by filling conversation ellipsis is closely related to previous research on ellipsis resolution and natural language understanding. Here, we choose two language understanding tasks: dialog act prediction and semantic role labeling, which can significantly influence the performance of deployable social chatbots. 
\subsection{Ellipsis Completion}
Automated ellipsis completion traditionally adopted rule-based methods \cite{dalrymple1991ellipsis}. There has been a long line of research on verb ellipsis recovery. \cite{hardt-1997-empirical} determined potential antecedents by applying syntactic constraints.  \cite{dienes-dubey-2003-deep,dienes-dubey-2003-antecedent} investigated antecedent recovery with a trace tagger. Later \cite{nielsen2003corpus} completed verb ellipsis in an end-to-end formulation. Then a verb phrase ellipsis detection system was designed using automatically parsed free text \cite{Nielsen:2004:VPE:1220355.1220512}. \cite{schuster-etal-2018-sentences} further studied methods of parsing to a Universal Dependencies graph representation to reconstruct predicates in sentences with gapping. However, only recently has research in ellipsis completion for dialog been published. \cite{DBLP:journals/corr/abs-1906-07004} proposed to recover ellipsis through utterance rewriting on Chinese conversations. Although our completion model is also built upon Pointer Generator \cite{vinyals2015pointer,DBLP:conf/acl/GuLLL16,see2017get}, we intend to improve downstream language understanding, whose accuracy has been highly emphasized by large-scale deployable social chatbots. 

\subsection{Dialog Act Prediction}
Dialog act prediction aims to classify the intention or function of a speaker's utterance (e.g. open question, statement). Previously, most deep learning neural models were trained and evaluated on human-human conversations. \cite{liu2017using} proposed to use a hierarchical RNN or CNN for dialog act sequence tagging. \cite{Chen:2018:DAR:3209978.3209997} extended a structured attention network to the linear-chain conditional random field layer.   \cite{raheja2019dialogue} further coupled the hierarchical RNN with a context-aware self-attention mechanism. Recently, \cite{yu2019midas} proposed a new dialog act annotation scheme, MIDAS, for open-domain human-machine conversations and proposed a multi-label dialog act prediction model leveraging pretrained BERT \cite{devlin2018bert} for utterance encoding. We follow the same model and focus on combining original user utterances with automatically completed utterances to further improve dialog act prediction in the context of human-machine conversation.

\subsection{Semantic Role Labeling}
Semantic role labeling (SRL) is a task of providing semantic relations between arguments and predicates. Traditional approaches to SRL employed linear classifiers based on hand-crafted feature templates \cite{Pradhan:2005:SRC:1706543.1706585,punyakanok-etal-2008-importance}. Recent approaches provide end-to-end deep models for SRL without syntactic information for input \cite{Zhou2015EndtoendLO}. The performance was further improved using deep highway Bi-LSTMs with constrained decoding \cite{he-etal-2017-deep}. \cite{tan2018deep} applied a self-attention mechanism on SRL to solve problems concerning memory compression and inherent sentence structure when using RNNs. Using features induced by neural networks, these models predict a ``BIO" tag for each token. Our work and model for SRL utilize this BIO tagging approach. However, we address the challenge of resolving prevalent ellipsis in social conversations that previous state-of-the-art models do not focus on. Further, we demonstrate that resolving ellipsis improves performance in downstream language understanding tasks like semantic role labeling.

\section{Proposed Model}
\subsection{Problem Formalization}
We aim to improve a series of language understanding tasks by combining utterances containing ellipsis with their auto-completed counterparts. Our formal problem definition is as follows. $T$ represents the set of language understanding tasks where $|T|=N$. $U_E$ represents the set of original input utterances with ellipsis and $U_C$ represents the set of utterances after completion. $L_i, i=1,...N$ represents the task-specific label space. For example, in the dialog act prediction task, $L_i$ is the set of predicted dialog acts. In the semantic role labeling task, $L_i$ is the set of predicted BIO sequence tags. We first learn an end-to-end completion model to automatically complete utterances with ellipsis. The model is represented as $f: U_E \rightarrow U_C$. Then for a specific task $t_i \in T,i=1,...N$, our goal is to predict $L_i$ based on $U_E$ and $U_C$.

\subsection{Hybrid-EL-CMP}
We present our framework Hybrid-ELipsis-CoMPlete (Hybrid-EL-CMP) that utilizes information from the utterances with ellipsis and their auto-completed counterparts. There are three components in Hybrid-EL-CMP: a completion model, two encoder-classifier models, and a selection model that considers information from both an utterance with ellipsis and after auto-completion. Figure \ref{fig:1} shows an overview of Hybrid-EL-CMP. Note that we use the dialog act prediction task as an example for illustration. The framework could easily be generalized to other dialog understanding tasks. We now provide additional details about the three primary system components.

\subsubsection{Completion Model}
We first train an end-to-end sequence-to-sequence model with copy mechanism to automatically complete utterances that contain ellipsis. 

Our completion model is based on the Pointer Generator \cite{see2017get} which is a combination of the vanilla Seq2Seq with attention \cite{bahdanau2014neural} and the pointer network \cite{vinyals2015pointer}. The Pointer Generator allows copying words directly from the context (previous user utterances in our case) while retaining the ability to generate words from the decoder.  These copied words are likely to be the omitted information that we want to complete. Here we use $\lambda$ to represent a switch probability between generation mode and copy mode. $\lambda$ is calculated from the encoder context vector $h^*$, the decoder input $x_t$ and the decoder hidden states $s_t$ at timestep $t$:
$$\lambda = sigmoid (W_\lambda[h^*,x_t,s_t]+b_\lambda)$$
where $W_\lambda$ and $b_\lambda$ are learnable parameters. We use $P_{gen} (w)$ to represent the distribution over the whole vocabulary to generate word $w$ from the decoder, $P_{copy} (w)$ to represent the distribution of copying words from the original context, and $a^t$ to represent the attention distribution. The final distribution to predict word $w$ is calculated as:
\begin{align}
P_{gen} (w)&=g (h^*,s_t)\\
P_{copy} (w)&=\sum_i a_i^t\\
   P (w)&=softmax ([\lambda P_{gen} (w), (1-\lambda)P_{copy} (w)])
\end{align}

\subsubsection{Language Understanding Encoder and Classifier} 
We apply two encoder-classifier models in Hybrid-EL-CMP as shown in Figure \ref{fig:1}. The model above is for encoding utterances with ellipsis and the model below is for encoding utterances after completion. For dialog act prediction, we leverage the BERT model trained on Wikipedia \cite{devlin2018bert}. For semantic role labeling, we leverage stacked Bi-LSTMs with highway connections trained on CONLL2012 \cite{pradhan-etal-conll-st-2012-ontonotes}, similar to \cite{he-etal-2017-deep}.

\subsubsection{Selection Model} We experiment with several selection methods to combine the information of utterances with ellipsis and utterances after completion. We divide these methods into two types: logits-based methods and hidden-states-based methods. 

For logits-based selection methods, we apply two classifiers after two encoders and get two distributions, $D_E$ (distribution of utterances with ellipsis) and
$D_C$ (distribution of utterances after completion) over our label space $L_i$. Our final distribution $D$ over $L_i$ can be formalized as the sum ($D_{sum}$) or the max ($D_{max}$) of the original two distributions.
\begin{align}
    D_{sum} &= D_E + D_C \\
    D_{max} &= max\{D_E,D_C\}
\end{align}

For hidden-states-based selection methods, we combine the information after two encoders and apply one classifier. Let $H_E$ denote the encoder hidden states of utterances with ellipsis and $H_C$ denote the encoder hidden states of utterances after completion. Our selection model can be formalized as the sum ($H_{sum}$), the max ($H_{max}$) or the concatenation ($H_{cat}$) of the original two encoder hidden states. In general, we denote these hidden state combinations as $H$.
\begin{align}
    H_{sum} &= H_E + H_C \\
    H_{max} &= max\{H_E,H_C\}\\
    H_{cat} &= [H_E | H_C]
\end{align}
The final distribution is calculated as:
\begin{equation}
    D=W*H+b
\end{equation}
where $W$ is the  weight matrix and $b$ is a bias vector.

\subsubsection{Dialog Act Prediction}
In our selection model, we also exploit some expert knowledge to further adapt to our specific dialog understanding task. Here we demonstrate how we incorporate this expert prior knowledge by taking dialog act prediction and semantic role labeling tasks as two examples. 
For dialog act prediction task, let $L_{DA}$ denote all possible dialog acts. Based on detailed review of our dialog act scheme, we define specific dialog acts that are not suitable to be predicted from completed utterances (eg. hold, complaint as shown in Table \ref{tab:elli_example}), denoted as $L_{DA_{non}} \subseteq L_{DA}$. If our model predicts such dialog acts in $L_{DA_{non}}$ from the original utterance, we directly use that prediction as the final output. Otherwise, we combine the predictions from the original utterance and the automatically completed utterance. For example, consider utterances with dialog act such as ``hold" (e.g. ``okay"; see detailed example in Table \ref{tab:elli_example}), which have higher error rates in the completion process. In this case, the user pauses before saying ``let's change conversation". The completion model might complete the utterance to be ``okay I have read any other books by the same author", which may confuse the dialog act prediction model that the dialog act of this utterance is ``positive answer".
Therefore, expert information can guide the selection module to make more accurate decision.

\subsubsection{Semantic Role Labeling}
For the semantic role labeling task, we have incorporated two kinds of expert knowledge: rule-based and probability-based knowledge. 
\begin{itemize}
    \item \textit{Rule-based expert knowledge}: If both original utterances and auto-completed utterances have predicates, SRL on the original utterances tends to provide more satisfying results than its completed counterparts due to possible auto-completion errors. For example from our Gunrock dataset, the user says ``I watch TV more than I watch movies", and the auto-completed utterance is ``I watch TV more than I watch", which misses ``movies" as ``ARG1".  Therefore, we design a rule-based selection method: If the original utterance has a predicate, then we just output the semantic roles from the original sentence; Otherwise, we first auto-complete the utterance and then perform semantic role labeling. 
    \item \textit{Probability-based expert knowledge}: If both the original utterance and the auto-completed utterance have predicates, though in general SRL on the original utterances is better, for a specific argument in the utterance, SRL from the auto-completed utterance could give better performance with some probability. This probability correlates to the beam search posterior probability for this argument in our completion model. Specifically, we first set a threshold. Then for a given argument, we consider each token comprising it. If any of these tokens has a beam search posterior probability less than the threshold, we regard that the auto-completion quality is not that good so we predict SRL according to the original utterance.
\end{itemize}

\section{Dataset and Annotation Scheme}
We evaluate our Hybrid-EL-CMP on a dataset collected in our in-lab user studies with a social bot on the Alexa platform (Gunrock dataset) \cite{chen2018gunrock}. This dataset provides real human-machine social conversations that cover a broad range of topics including sports, politics, entertainment, technology, etc. We use five-fold cross validation to conduct hyperparameter tuning of our models. Once we have identified the optimal hyperparameters, such as number of epochs and learning rate, we combine the validation and training data for final model training. Finally, we report results on a held-out test dataset.

\subsection{Utterance Completion Scheme}
We design an utterance completion scheme as follows:
\begin{itemize}
    \item If the original utterance has ellipsis, then we manually complete the utterance given context information.
    \item If the original utterance is complete and may be readily modified to create an example of ellipsis
    , then we modify the utterance to create a version containing ellipsis.
    \item If the utterance is complete and not appropriate for creating an ellipsis version, we just keep the original utterance.
\end{itemize}
We randomly selected 2,258 user utterances from the Gunrock dataset for utterance completion. Among them 1,124 utterances have ellipsis, and 204 utterances are complete but can be modified to a version with ellipsis. The rest are complete and cannot be modified for ellipsis.

\subsection{Dialog Act Annotation Scheme}
We follow the scheme of MIDAS \cite{yu2019midas} for dialog act prediction. In total we have 11,602 user utterances with 23 dialog acts. There are two main types of dialog act: semantic requests and functional requests. Semantic requests capture dialog content information such as open question, command, statement, etc. Functional requests help improve discourse coherence and are composed of incomplete, social convention and other classes, such as nonsense, apology, opening, etc.

\subsection{Semantic Role Labeling Scheme}
For Semantic role labeling, we randomly chose 1,689 user utterances from the same Gunrock dataset, of which 21.73\% contain verb ellipsis. We follow OntoNotes 5.0 \cite{ontonote5.0} to annotate semantic roles. OntoNotes is a span-based annotation scheme which was originally designed for formal text. However, dialog utterances with ellipsis may not have explicit predicates. Therefore, we make several modifications to the original annotation scheme to adapt it to dialog settings.

\begin{itemize}
    \item If a user utterance contains no predicate, it will be annotated using the predicate in the interlocutor's previous utterance as shown in Table \ref{tab:srl_anno_example1}.
\begin{table*}[htbp]
        \begin{tabular}{c|c|c|c}
        \hline
            
            \textbf{Case}&\textbf{System}&\textbf{User}&\textbf{SRL} \\
        \hline
            \multirow{2}*{1} & 
            \multirow{2}{50mm}{what do you want to talk about}&
            \multirow{2}{32mm}{guitars} &
            \multirow{2}{68mm}{(talk about)[ARG1:guitars]} \\
            & & & \\
        \hline
            \multirow{2}*{2} &
            \multirow{2}{50mm}{speaking of which , how often do you play it}&
            \multirow{2}{32mm}{every single day}&
            \multirow{2}{68mm}{(play it)[ARGM-TMP:every single day]} \\
            & & &\\
        \hline
              \multirow{2}*{3} &
              \multirow{2}{50mm}{do you prefer to watch movies in the theater or at home}&
              \multirow{2}{32mm}{at home}&
              \multirow{2}{68mm}{(watch movies)[ARGM-LOC:at home]} \\
              & & &\\
        \hline
        \end{tabular}
        
        \caption{Examples of new annotation scheme for utterances with no predicates. The content in brackets in SRL column is the interlocutor's previous utterance, according to which we annotate the incomplete utterance from user.}
        \label{tab:srl_anno_example1}
    \end{table*}
    \item If a user utterance is a subordinate clause, it will be annotated according to the relativizer in the previous system output. For example, the entire utterance will be specifically annotated as an object if it is an object clause. This only influences the a few layers of SRL prediction and other predicates in it (if they exist) will form their own predicting layer normally as shown in Table \ref{tab:srl_anno_example2}.
    
\begin{table*}[htbp]
        \begin{tabular}{c|c|c|c}
        \hline
            
           \textbf{Case}&\textbf{System}&\textbf{User}&\textbf{SRL}  \\
        \hline
            \multirow{2}*{1} &
            \multirow{2}{50mm}{what part did you like best about that movie}&
            \multirow{2}{32mm}{when the robots did fight}&
            \multirow{2}{68mm}{ (the part)[ARG1:when the robots did fight]}\\
            & & &\\
         \hline
            \multirow{2}*{2}& 
            \multirow{2}{50mm}{do you enjoy traveling}&
            \multirow{2}{32mm}{when I was younger}&
            \multirow{2}{68mm}{(enjoy traveling)[ARGM-TMP:when I was younger]}\\
            & & &\\
        \hline
        \end{tabular}
        
        \caption{Examples of new annotation scheme for utterances that are subordinate clauses. The content in brackets in SRL column is the previous system output, according to which we annotate the incomplete utterance from user.}
        \label{tab:srl_anno_example2}
    \end{table*}
\end{itemize}

\section{Utterance Completion Experiments}

\subsection{Experimental Settings}
Using our annotated utterance completion dataset, we first train the automatic completion model. We compare two Seq2Seq utterance completion models, one with a copy mechanism and one without. For both models, the encoder and decoder are 2-layer LSTMs and we set the hidden state size to 500. The dropout rate is 0.3. We train the models leveraging OpenNMT \cite{klein-etal-2017-opennmt} with an SGD optimizer. The initial learning rate is 1.
\subsection{Experimental Results}
\begin{table}[ht]

    \resizebox{\columnwidth}{!}{
    \begin{tabular}{cccccc}
    \hline
    \textbf{Model}&\textbf{BLEU}(\%)&\textbf{EM}(\%)&\textbf{Prec.}(\%)&\textbf{Rec.}(\%)&\textbf{F1}(\%)\\
    \hline
    \hline
        Seq2Seq&42.36&28.50&61.28&61.25&61.13\\
 
        Seq2Seq+Copy&\textbf{71.85}&\textbf{59.81}&\textbf{89.46}&\textbf{89.22}&\textbf{89.28}\\
    \hline
    \end{tabular}}

    \caption{Automatic completion results. EM represents exact match rate.}
    \label{tab:resolution_task}
\end{table}

 We show the performance of models with and without a copy mechanism in Table \ref{tab:resolution_task}. We compare the two models in terms of BLEU, EM, one word precision, recall and F1 score. We observe a huge performance gain by incorporating the copy mechanism. 

\subsection{Case Study}
We further analyze the strengths and weaknesses of our copy-based Seq2Seq model. Our completion model performs well in the following three cases: (1) If the original utterance is already complete in itself, then the completion model can learn to copy the utterance and does not disturb or miss the original information. For example, the system says ``sadly, I can only look at animal videos online" and the user asks ``how can you see if you don't have eyes". In this case, the user utterance is already complete and our auto-completed utterance is the same as the original utterance. (2) If the user responds directly to the system's question, our completion model can correctly find the omitted information. For example, the system asks ``what is your favorite movie" and the user replies ``titanic". In this case, the completion model can complete the utterance correctly to generate ``My favorite movie is titanic." (3) If the user proposes a new topic, our completion model can also infer from the context and resolve the missing utterance. For example, the system asks ``do you want to talk about football" and the user proposes ``how about movies". In this case, our model can complete the utterance to be ``how about talking about movies". 

We categorize common completion errors into the following three situations: (1) The model might complete some utterances that should not be completed as shown in case 2 of Table \ref{tab:elli_example}.  To handle such errors, we include expert knowledge in the dialog act prediction model to set utterances with certain predicted dialog acts to not to be completed.
(2) There are paraphrases. For example, the system asks ``would you like to keep talking about technology?" and the user says ``yes technology". While the ground truth completed utterance is ``yes I would like to keep talking about technology", our model might complete it to be ``yes I want to keep talking about technology". Although the auto-completed utterance is not exactly the same as the ground truth, it does not change the predicted dialog act. (3) There are minor words missing or repetition. For example, the system asks ``do you think it would be true through their whole life", and the user answers ``yes." The auto-completed utterance is ``yes I think would would be true through their whole life" which repeats the word ``would".

\section{Dialog Act Prediction Experiments}
\subsection{Experimental Settings}
Once we obtain automatically completed utterances, we perform dialog act prediction using our proposed framework,  Hybrid-EL-CMP. We compare the model with four baselines:
\begin{itemize}
    \item EL: a single model trained on utterances with ellipsis
    \item CMP: a single model trained on utterances after completion
    \item Hybrid-EL-EL: two models both trained on utterances with ellipsis
    \item Hybrid-CMP-CMP: two models both trained on utterances after completion
\end{itemize}
We leverage pretrained BERT for all the encoders and adopt the same evaluation metrics as the state-of-the-art dialog act prediction model \cite{yu2019midas}. We train the model with the Adam optimizer. The initial learning rate is 5e-5. 
\subsection{Experimental Results}
Our model Hybrid-EL-CMP proves to outperform all the baselines. Dialog act prediction results are summarized in Table \ref{tab:DA_task}. We observe that Hybrid-EL-EL, combining results from two models which both use original sentences with ellipsis, slightly outperforms EL which has a single model utilizing sentences with ellipsis. Similarly we find Hybrid-CMP-CMP slightly outperforms CMP. This is because ensemble models generally perform better than a single model.  
Moreover, a model using only the auto-completed sentence does not perform as well as a model using the original sentence with ellipsis. This is because automatic completion errors can carry over. However, jointly utilizing utterances with ellipsis and utterances after completion, our Hybrid-EL-CMP reaches the best results in terms of precision, recall and F1 score.
\begin{table}[ht]
\centering
    \begin{tabular}{cccc}
    \hline
   
        \textbf{Model}&\textbf{Prec.}(\%)&\textbf{Rec.}(\%)&\textbf{F1}(\%)\\
    \hline
    \hline
        EL&80.32&79.80&79.65\\
  
        CMP&79.06&77.92&78.04\\
   
        Hybrid-EL-EL&80.37&79.91&79.70\\
  
        Hybrid-CMP-CMP&79.43&78.95&78.74\\
  
        Hybrid-EL-CMP&\textbf{81.30}&\textbf{81.41}&\textbf{80.90}\\
    \hline
    \end{tabular}
    \caption{Hybrid-EL-CMP performs the best in dialog act prediction.}
    \label{tab:DA_task}
\end{table}

 We also study the effects of different selection methods in our selection model. We can see from Table \ref{tab:Selection} that empirically adding logits from two models after classifiers performs the best. Besides, adding information of two models generally performs better than methods of concatenation or finding the maximum, whether addition is implemented on hidden states after encoders or logits after classifiers.  In addition, incorporating expert knowledge can further improve performance. Here the expert knowledge is added only during testing. We have also tried adding expert knowledge both during training and testing by applying tensor masks on logits from two models according to our pre-defined set of dialog acts not to be completed. We find that incorporating expert knowledge only during testing empirically performs better.\\
\begin{table}[ht]
    \begin{tabular}{cccc}
    \hline
      
        \textbf{Selection Method}&\textbf{Prec.}(\%)&\textbf{Rec.}(\%)&\textbf{F1}(\%)\\
    \hline
    \hline
        Max Logits&80.19&80.50&79.85\\
  
        Add Logits&81.30&81.28&80.85\\
        Add Logits+Expert&\textbf{81.30}&\textbf{81.41}&\textbf{80.90}\\ 
    \hline
        Concat Hidden&80.24&80.04&79.65\\
  
        Max Hidden&80.30&80.04&79.63\\
  
        Add Hidden&80.82&80.28&80.08\\
  
    \hline
    \end{tabular}
    \caption{Dialog act prediction performance using different selection methods.}
    \label{tab:Selection}
\end{table}
\subsection{Case Study}
We provide several common cases to illustrate that our Hybrid-EL-CMP can improve dialog act prediction. All examples are shown in Table \ref{tab:DA_case_study}.

\begin{table*}
\centering
    {
    \resizebox{2\columnwidth}{!}
    {
    \begin{tabular}{c|c|c|c|c|c|c|c}
    \hline
        
        \multirow{3}{*}{\textbf{Case}}&
        \multirow{3}{*}{\centering{\textbf{System}}}&
        \multirow{3}{20mm}{\centering{\textbf{User\\Ellipsis}}}&
        \multirow{3}{20mm}{\centering{\textbf{User\\Complete}}}&
        \multirow{3}{11mm}{\textbf{Act\\Ellipsis}}&
        \multirow{3}{15mm}{\textbf{Act\\Complete}}&
        \multirow{3}{17mm}{\textbf{Act Ellipsis\\+Complete}}&
        \multirow{3}{17mm}{\textbf{Act Ground\\Truth}}\\
        & & & & & & &\\
        & & & & & & &\\
    \hline
    
    \multirow{4}*{1}&
    \multirow{4}{30mm}{If music were removed from the world, how would you feel?}&
    \multirow{4}{20mm}{Sad.}&
    \multirow{4}{25mm}{{\textit{I would feel}} sad.}& \multirow{4}*{comment}& \multirow{4}*{opinion}& 
    \multirow{4}*{opinion}&
    \multirow{4}*{opinion}\\
    & & & & & & &\\
    & & & & & & &\\
    & & & & & & &\\
    \hline
    
    \multirow{5}*{2}&
    \multirow{5}{30mm}{If you got the chance to ask the director of that movie one question, what would it be? }&
    \multirow{5}{20mm}{What's up with that scene at the end.}  & 
    \multirow{5}{25mm}{{\textit{I would ask}} what's up with that scene at the end.}&
    \multirow{5}{12mm}{\centering{open\\factual\\question}}&
    \multirow{5}*{opinion}&
    \multirow{5}*{opinion}&
    \multirow{5}*{opinion}\\
    & & & & & & &\\
    & & & & & & &\\
    & & & & & & &\\
    & & & & & & &\\
    \hline
    \multirow{4}*{3}&
    \multirow{4}{30mm}{Have you read any other books by the same author?}&
    \multirow{4}{20mm}{Okay. (Let's change conversation.)}&
    \multirow{4}{25mm}{Okay {\textit{I have read any other books by the same author.}}}&
    \multirow{4}*{hold}&
    \multirow{4}{11mm}{\centering{positive\\answer}}&
    \multirow{4}*{hold}&\multirow{4}*{hold}\\
    & & & & & & &\\
    & & & & & & &\\
    & & & & & & &\\
    \hline

    \multirow{3}*{4}&
    \multirow{3}{30mm}{{Have you ever had a pet?}}&
    \multirow{3}{20mm}{Yes I have a pet.}&
    \multirow{3}{25mm}{Yes I have a pet.}&
    {positive}&\multirow{3}{11mm}{\centering{positive\\answer}}&
    {positive} &{positive}\\
    & & & &answer;& &answer;&answer;\\
    & & & &statement & &statement &statement\\

    \hline
    \end{tabular}}}
    \caption{Four examples of dialog act prediction task. The first two lines show cases when original utterances predict the incorrect dialog acts while auto-complete utterances predict correct dialog acts. The last two lines are reversed. In all four cases, our Hybrid-EL-CMP predicts the correct dialog acts. Italics represents the automatically completed part.}
    \label{tab:DA_case_study}
\end{table*}

The first and second example show how completion resolves ambiguity. In case 1, the utterance contains a short statement of opinion. For most other cases, short responses are comments like ``great" and here the system mistakenly recognizes ``sad" as comment. However, if we complete the original utterance to be ``I would feel sad", the system will correctly identify that the user is expressing his or her feeling. Case 2 shown in Table \ref{tab:elli_example} also suggests completion resolves ambiguity of the utterance.

The third and fourth examples demonstrate that completion sometimes introduces other kinds of misunderstandings. To tackle problems like case 3, we have incorporated a predefined set of dialog acts not to be completed as illustrated in the model section. The fourth example shows that Hybrid-EL-CMP overcomes the problem that automatic completion might obscure some responses. As the ground truth dialog acts are annotated on the original utterances, corresponding dialog acts of the same auto-completed utterances might be different. For instance, whether the user responds with a simple "yes" or "yes I have had a pet", after automatic completion the response would be the same as "yes I have had a pet". But their corresponding dialog acts are different, i.e. positive answer and positive answer;statement, respectively. If we train with the automatically completed utterances, we cannot accurately disambiguate this difference in the original dialog act. However, by combining original utterances, we can address this problem and correctly identify the dialog act as positive answer;statement.

\section{Semantic Role Labeling Experiments}

\subsection{Experimental Settings}
Apart from dialog act prediction, we also evaluate our proposed Hybrid-EL-CMP on semantic role labeling task. We leverage stacked Bi-LSTMs similar to \cite{he-etal-2017-deep}. We make two changes to the annotation scheme because of ellipsis cases are not properly considered under it and thus we adjust the evaluation metrics by the following: 
\begin{itemize}
    \item The empty output will be ignored under the original annotation scheme. Thus the false-negative score can't reflect the actual performance properly. Now, the empty output will not be ignored. Instead it will be added to false-negative score for penalty according to ground-truth labels.
    \item Because of auto-completion, the completed part is also predicted SRL labels. We only compare the labels of the original utterance
    So instead of evaluating the whole output of completion sentences, only the labels of the corresponding parts contribute to the evaluation metrics.
\end{itemize}

\subsection{Experimental Results}
\begin{table}[ht]
    \centering
    \begin{tabular}{cccc}
    \hline
      
        \textbf{Model}&\textbf{Prec.}(\%)&\textbf{Rec.}(\%)&\textbf{F1}(\%)\\
    \hline
    \hline
        EL&96.02&81.89&88.39\\
  
        CMP&86.39&\textbf{88.64}&87.50\\
    \hline
        Hybrid-EL-CMP1&\textbf{97.42}&84.70&90.62\\
        Hybrid-EL-CMP2&95.82&86.42&\textbf{90.87}\\
    \hline
    \end{tabular}
    \caption{Semantic role labeling results. Hybrid-EL-CMP1 represents rule-based model and Hybrid-EL-CMP2 represents probability-based model.}
    \label{tab:SRL}
\end{table}

Our results in Table \ref{tab:SRL} show that when only using original utterances with ellipsis, precision is relatively high while recall is low. This indicates that original utterances give more precise labels if verb ellipsis does not occur. On the other hand, the result of only using auto-completed utterances is reversed, with low precision and high recall. This also conforms to our assumption that our completion model sometimes makes grammatical errors or simply misses something from the original utterance. So the precision is relatively low. However, auto-completion produces predicates for verb ellipsis cases and subordinate-clause cases so that more labels could be predicted. Thus the recall of using utterances after completion is higher. By combining the advantages of original utterances and utterances after completion, our Hybrid-EL-CMP gives the best F1 score. We further analyze our two hybrid models guided by two kinds of expert knowledge illustrated in the model section. The probability-based knowledge performs better when the entity name contains verbs. For example, the user states the name of his favorite game ``detroit become human". While the entire entity name is ``ARG1", the rule-based knowledge simply chooses the wrong SRL prediction from the ellipsis utterances. So the recall of the rule-based knowledge is lower than that of probability-based knowledge. On the other hand, in cases where completion makes small mistakes, such as completing the name ``david" to be ``david david", the beam search posterior probability of these tokens is still quite high. So the model mistakenly chooses the results from auto-completed utterances, resulting in a relatively low precision. 

\subsection{Case Study}
\begin{figure}
\centering
\includegraphics[width=1.0 \columnwidth]{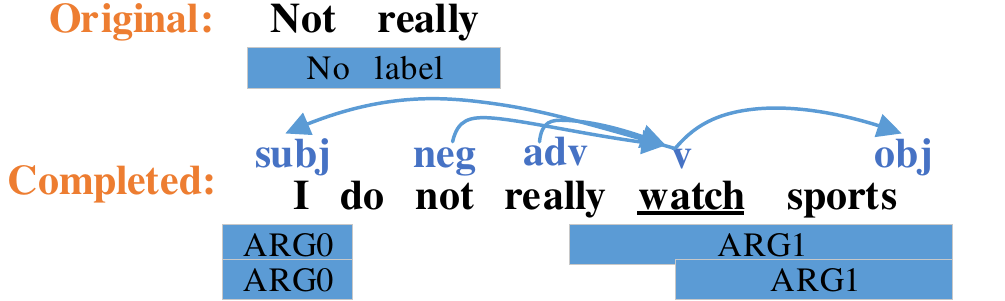}
\caption{Example of completion outperforming ellipsis. The blue boxes contain labeling result.}
\label{fig:2}
\end{figure}

\begin{figure}
\centering
\includegraphics[width=1.0 \columnwidth]{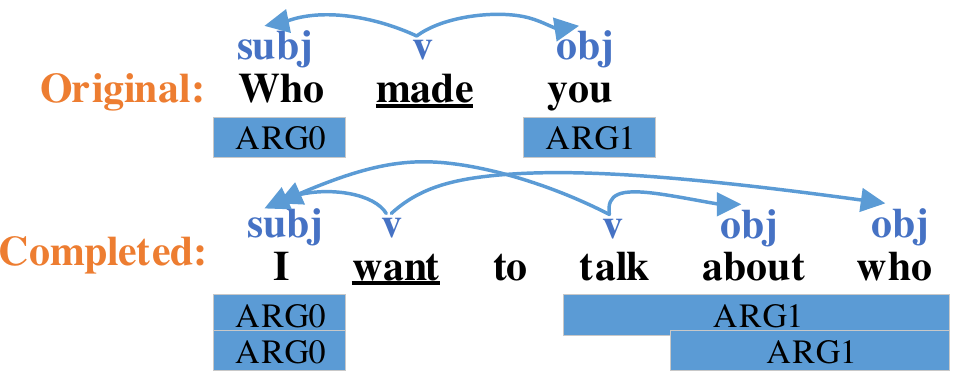}
\caption{Example of ellipsis outperforming completion. The blue boxes contain labeling result.}
\label{fig:3}
\end{figure}

Figure \ref{fig:2} shows a case where the user utterance contains verb ellipsis. Here, the ``not really" utterance from the user does not contain a verb. But the sentence is sufficient to express whether the user watches sports or not. Traditional SRL model requires a verb function. The missing verb and argument lie in the previous utterance: ``do you watch sports". 
Therefore, when an utterance has verb ellipsis, the model primarily relies on the SRL output of the automatically completed sentence.\\

Figure \ref{fig:3} shows a case when the original utterance contains a predicate but the auto-completed utterance lacks necessary semantic information and produces a comparatively different sentence. Here, the utterance from the dialog system is ``what do you want to talk about". Instead of responding to the question posed by the system, the user proposes a new topic by asking a question directly. Auto-completion erroneously completes the user utterance to be ``I want to talk about who", which leads to SRL errors. Our hybrid model will utilize the SRL prediction of the ellipsis utterance and provide correct output.

\section{Conclusion}
Ellipsis frequently occurs in social conversations. We propose Hybrid-ELlipsis-CoMPlete (Hybrid-EL-CMP) to utilize both original utterances with ellipsis and their automatically completed counterparts for improving language understanding in human-machine social conversations. We show the effectiveness of the proposed approach on language understanding tasks by evaluating on dialog act prediction and semantic role labeling. 
We believe that the framework can be generalized to other dialog understanding tasks as well, such as syntactic and semantic parsing. We will evaluate these tasks and also our proposed model on other domains such as human-human conversations in the future.

\bibliography{reference}

\begin{thebibliography}{}

\bibitem[\protect\citeauthoryear{Bahdanau, Cho, and
  Bengio}{2014}]{bahdanau2014neural}
Bahdanau, D.; Cho, K.; and Bengio, Y.
\newblock 2014.
\newblock Neural machine translation by jointly learning to align and
  translate.
\newblock {\em arXiv preprint arXiv:1409.0473}.

\bibitem[\protect\citeauthoryear{Chen \bgroup et al\mbox.\egroup
  }{2018a}]{chen2018gunrock}
Chen, C.-Y.; Yu, D.; Wen, W.; Yang, Y.~M.; Zhang, J.; Zhou, M.; Jesse, K.;
  Chau, A.; Bhowmick, A.; Iyer, S.; et~al.
\newblock 2018a.
\newblock Gunrock: Building a human-like social bot by leveraging large scale
  real user data.

\bibitem[\protect\citeauthoryear{Chen \bgroup et al\mbox.\egroup
  }{2018b}]{Chen:2018:DAR:3209978.3209997}
Chen, Z.; Yang, R.; Zhao, Z.; Cai, D.; and He, X.
\newblock 2018b.
\newblock Dialogue act recognition via crf-attentive structured network.
\newblock In {\em The 41st International ACM SIGIR Conference on Research
  \&\#38; Development in Information Retrieval}, SIGIR '18,  225--234.
\newblock New York, NY, USA: ACM.

\bibitem[\protect\citeauthoryear{Dalrymple, Shieber, and
  Pereira}{1991}]{dalrymple1991ellipsis}
Dalrymple, M.; Shieber, S.~M.; and Pereira, F.~C.
\newblock 1991.
\newblock Ellipsis and higher-order unification.
\newblock {\em Linguistics and philosophy} 14(4):399--452.

\bibitem[\protect\citeauthoryear{Devlin \bgroup et al\mbox.\egroup
  }{2018}]{devlin2018bert}
Devlin, J.; Chang, M.-W.; Lee, K.; and Toutanova, K.
\newblock 2018.
\newblock Bert: Pre-training of deep bidirectional transformers for language
  understanding.
\newblock {\em arXiv preprint arXiv:1810.04805}.

\bibitem[\protect\citeauthoryear{Dienes and
  Dubey}{2003a}]{dienes-dubey-2003-antecedent}
Dienes, P., and Dubey, A.
\newblock 2003a.
\newblock Antecedent recovery: Experiments with a trace tagger.
\newblock In {\em Proceedings of the 2003 Conference on Empirical Methods in
  Natural Language Processing},  33--40.

\bibitem[\protect\citeauthoryear{Dienes and
  Dubey}{2003b}]{dienes-dubey-2003-deep}
Dienes, P., and Dubey, A.
\newblock 2003b.
\newblock Deep syntactic processing by combining shallow methods.
\newblock In {\em Proceedings of the 41st Annual Meeting of the Association for
  Computational Linguistics},  431--438.
\newblock Sapporo, Japan: Association for Computational Linguistics.

\bibitem[\protect\citeauthoryear{Gu \bgroup et al\mbox.\egroup
  }{2016}]{DBLP:conf/acl/GuLLL16}
Gu, J.; Lu, Z.; Li, H.; and Li, V. O.~K.
\newblock 2016.
\newblock Incorporating copying mechanism in sequence-to-sequence learning.
\newblock In {\em {ACL} {(1)}}.
\newblock The Association for Computer Linguistics.

\bibitem[\protect\citeauthoryear{Hardt}{1997}]{hardt-1997-empirical}
Hardt, D.
\newblock 1997.
\newblock An empirical approach to {VP} ellipsis.
\newblock {\em Computational Linguistics} 23(4):525--541.

\bibitem[\protect\citeauthoryear{He \bgroup et al\mbox.\egroup
  }{2017}]{he-etal-2017-deep}
He, L.; Lee, K.; Lewis, M.; and Zettlemoyer, L.
\newblock 2017.
\newblock Deep semantic role labeling: What works and what{'}s next.
\newblock In {\em Proceedings of the 55th Annual Meeting of the Association for
  Computational Linguistics (Volume 1: Long Papers)},  473--483.
\newblock Vancouver, Canada: Association for Computational Linguistics.

\bibitem[\protect\citeauthoryear{Klein \bgroup et al\mbox.\egroup
  }{2017}]{klein-etal-2017-opennmt}
Klein, G.; Kim, Y.; Deng, Y.; Senellart, J.; and Rush, A.
\newblock 2017.
\newblock {O}pen{NMT}: Open-source toolkit for neural machine translation.
\newblock In {\em Proceedings of {ACL} 2017, System Demonstrations},  67--72.
\newblock Vancouver, Canada: Association for Computational Linguistics.

\bibitem[\protect\citeauthoryear{Liu \bgroup et al\mbox.\egroup
  }{2017}]{liu2017using}
Liu, Y.; Han, K.; Tan, Z.; and Lei, Y.
\newblock 2017.
\newblock Using context information for dialog act classification in dnn
  framework.
\newblock In {\em Proceedings of the 2017 Conference on Empirical Methods in
  Natural Language Processing},  2170--2178.

\bibitem[\protect\citeauthoryear{Nielsen}{2003}]{nielsen2003corpus}
Nielsen, L.~A.
\newblock 2003.
\newblock A corpus-based study of verb phrase ellipsis.
\newblock In {\em Proceedings of the 6th Annual cluk Research Colloquium},
  109--115.

\bibitem[\protect\citeauthoryear{Nielsen}{2004}]{Nielsen:2004:VPE:1220355.1220512}
Nielsen, L.~A.
\newblock 2004.
\newblock Verb phrase ellipsis detection using automatically parsed text.
\newblock In {\em Proceedings of the 20th International Conference on
  Computational Linguistics}, COLING '04.
\newblock Stroudsburg, PA, USA: Association for Computational Linguistics.

\bibitem[\protect\citeauthoryear{Pradhan \bgroup et al\mbox.\egroup
  }{2005}]{Pradhan:2005:SRC:1706543.1706585}
Pradhan, S.; Hacioglu, K.; Ward, W.; Martin, J.~H.; and Jurafsky, D.
\newblock 2005.
\newblock Semantic role chunking combining complementary syntactic views.
\newblock In {\em Proceedings of the Ninth Conference on Computational Natural
  Language Learning}, CONLL '05,  217--220.
\newblock Stroudsburg, PA, USA: Association for Computational Linguistics.

\bibitem[\protect\citeauthoryear{Pradhan \bgroup et al\mbox.\egroup
  }{2012}]{pradhan-etal-conll-st-2012-ontonotes}
Pradhan, S.; Moschitti, A.; Xue, N.; Uryupina, O.; and Zhang, Y.
\newblock 2012.
\newblock {CoNLL-2012} shared task: Modeling multilingual unrestricted
  coreference in {OntoNotes}.
\newblock In {\em {Proceedings of the Sixteenth Conference on Computational
  Natural Language Learning (CoNLL 2012)}}.

\bibitem[\protect\citeauthoryear{Punyakanok, Roth, and
  Yih}{2008}]{punyakanok-etal-2008-importance}
Punyakanok, V.; Roth, D.; and Yih, W.-t.
\newblock 2008.
\newblock The importance of syntactic parsing and inference in semantic role
  labeling.
\newblock {\em Computational Linguistics} 34(2):257--287.

\bibitem[\protect\citeauthoryear{Raheja and
  Tetreault}{2019}]{raheja2019dialogue}
Raheja, V., and Tetreault, J.
\newblock 2019.
\newblock Dialogue act classification with context-aware self-attention.
\newblock {\em arXiv preprint arXiv:1904.02594}.

\bibitem[\protect\citeauthoryear{Schuster, Nivre, and
  Manning}{2018}]{schuster-etal-2018-sentences}
Schuster, S.; Nivre, J.; and Manning, C.~D.
\newblock 2018.
\newblock Sentences with gapping: Parsing and reconstructing elided predicates.
\newblock In {\em Proceedings of the 2018 Conference of the North {A}merican
  Chapter of the Association for Computational Linguistics: Human Language
  Technologies, Volume 1 (Long Papers)},  1156--1168.
\newblock New Orleans, Louisiana: Association for Computational Linguistics.

\bibitem[\protect\citeauthoryear{See, Liu, and Manning}{2017}]{see2017get}
See, A.; Liu, P.~J.; and Manning, C.~D.
\newblock 2017.
\newblock Get to the point: Summarization with pointer-generator networks.
\newblock {\em arXiv preprint arXiv:1704.04368}.

\bibitem[\protect\citeauthoryear{Su \bgroup et al\mbox.\egroup
  }{2019}]{DBLP:journals/corr/abs-1906-07004}
Su, H.; Shen, X.; Zhang, R.; Sun, F.; Hu, P.; Niu, C.; and Zhou, J.
\newblock 2019.
\newblock Improving multi-turn dialogue modelling with utterance rewriter.
\newblock {\em CoRR} abs/1906.07004.

\bibitem[\protect\citeauthoryear{Tan \bgroup et al\mbox.\egroup
  }{2018}]{tan2018deep}
Tan, Z.; Wang, M.; Xie, J.; Chen, Y.; and Shi, X.
\newblock 2018.
\newblock Deep semantic role labeling with self-attention.
\newblock In {\em AAAI Conference on Artificial Intelligence}.

\bibitem[\protect\citeauthoryear{Vinyals, Fortunato, and
  Jaitly}{2015}]{vinyals2015pointer}
Vinyals, O.; Fortunato, M.; and Jaitly, N.
\newblock 2015.
\newblock Pointer networks.
\newblock In {\em Advances in Neural Information Processing Systems},
  2692--2700.

\bibitem[\protect\citeauthoryear{Weischedel \bgroup et al\mbox.\egroup
  }{2013}]{ontonote5.0}
Weischedel; Palmer; Marcus; Hovy; Pradhan; Ramshaw; Nianwen-Xue; Taylor;
  Kaufman; Franchini; El-Bachouti; Belvin; Houston; et~al.
\newblock 2013.
\newblock Ontonotes release 5.0.

\bibitem[\protect\citeauthoryear{Yu and Yu}{2019}]{yu2019midas}
Yu, D., and Yu, Z.
\newblock 2019.
\newblock Midas: A dialog act annotation scheme for open domain human machine
  spoken conversations.
\newblock {\em arXiv preprint arXiv:1908.10023}.

\bibitem[\protect\citeauthoryear{Zhou and Xu}{2015}]{Zhou2015EndtoendLO}
Zhou, J., and Xu, W.
\newblock 2015.
\newblock End-to-end learning of semantic role labeling using recurrent neural
  networks.
\newblock In {\em ACL}.

\end{thebibliography}

\end{document}